\documentclass{article}
\usepackage{amssymb}
\usepackage{graphicx}
\usepackage[final]{corl_2026} 
\usepackage{subcaption} 
\usepackage{booktabs} 

\title{Implicit Neural Representations of Individual Behavior}

\author{
  Andrew Kang\\
  Carnegie Mellon University \\
  Electrical and Computer Engineering\\
  \texttt{akang2@andrew.cmu.edu} \\
  \And
  Priya Narasimhan\\
  Carnegie Mellon University \\
  Electrical and Computer Engineering\\  
  \texttt{priyan@andrew.cmu.edu} \\
}

\begin{document}
\maketitle


\begin{abstract}
We study policy representation learning from unlabeled multi-policy behavioral data. Each episode is generated by a fixed policy, but policy labels are unavailable. This setting appears in robotics play, demonstrations, games, racing, and other datasets where heterogeneous behaviors are mixed without annotations. We introduce \emph{Behavioral INR}, a self-supervised generative model that adapts implicit neural representations (INRs) from vision to behavior. Instead of mapping coordinates to RGB values, Behavioral INR represents a policy as a state-action function mapping states to subsequent actions. An episode-level latent modulates this function through FiLM layers, yielding a generative prior over policies and allowing policy identity to be inferred without supervision. Because INRs treat each datapoint as samples from an underlying function, the same model naturally accommodates variable episode lengths and different sampling granularities, as in vision INRs with different image resolutions. We also define policy-level out-of-distribution (OOD) shifts along state-distribution and action-distribution axes, which arise when policies overlap in states or actions but are not captured by standard behavioral OOD settings based only on new agents or environments. We evaluate on synthetic Gaussian random field data, MuJoCo demonstrations with controlled OOD splits, and real-world chess, Formula 1 racing, robotics, and Seek-Avoid datasets. Behavioral INR most consistently improves policy identifiability in the hardest continuous state-action settings, especially when longer episodes, more policies, and OOD splits reduce the usefulness of marginal shortcuts; amortized history encoders remain competitive when policy identity can be recovered from symbolic repetition or low-dimensional action statistics.
\end{abstract}



\section{Introduction}

\begin{figure*}[t]
    \centering
    \includegraphics[width=0.8\linewidth, height=8cm, keepaspectratio=false]{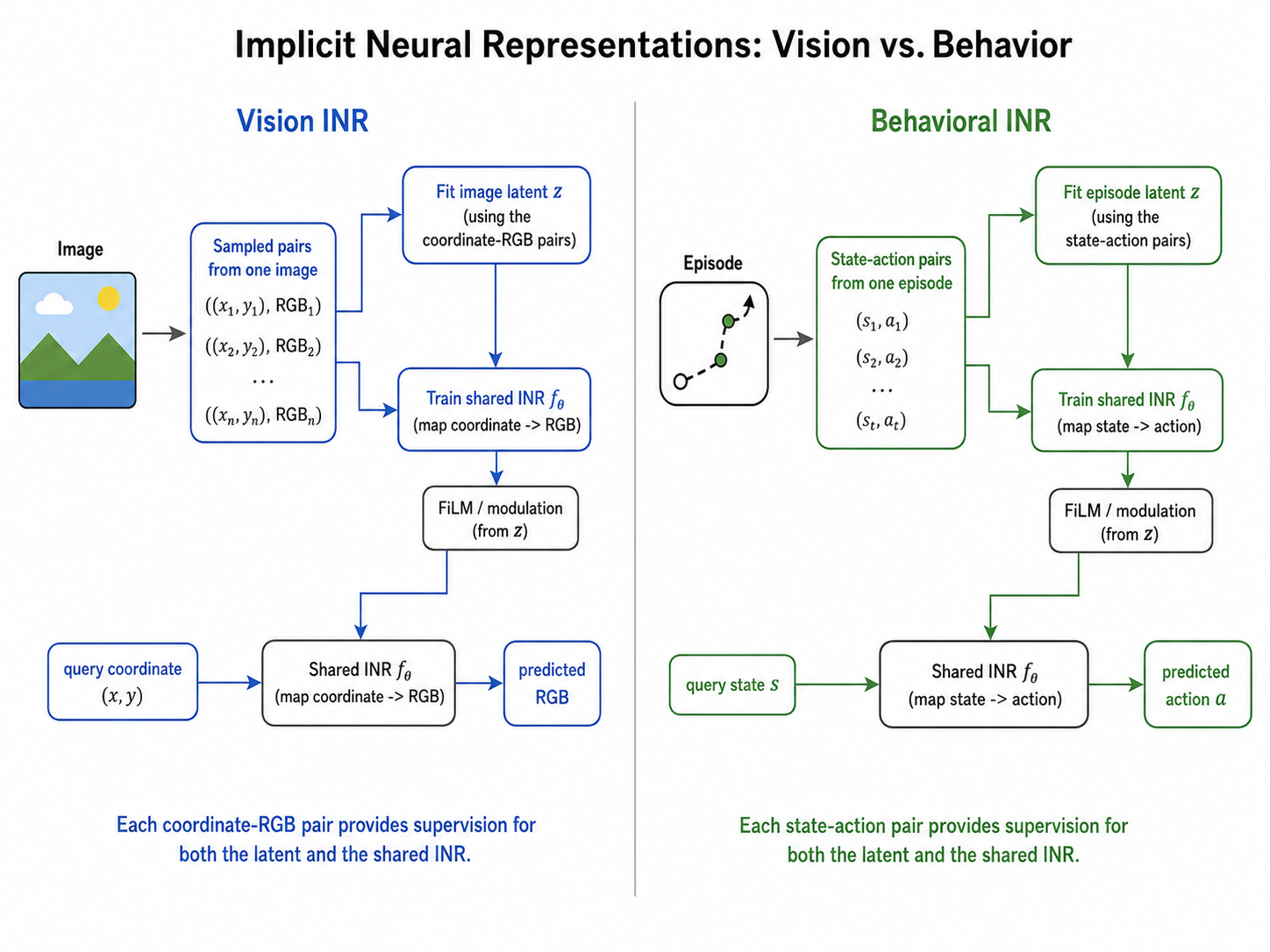}
    \caption{
    States and actions have the same relationship that pixel coordinates and RGB values have in implicit neural representations (INRs) for vision. Although behavior can be stochastic, partially observed, or history-dependent, existing imitation-learning models make the same state-to-action modeling abstraction. We find that this improves on previous work that relies on naive state-action history conditioning by concatenation. 
    }
    \label{fig:inr_figure}
\end{figure*}

Behavioral datasets in the wild often contain trajectories from many policies but do not identify which policy produced each episode. Recovering this latent policy identity is useful for understanding heterogeneous datasets, filtering training data, searching policy space, estimating matchups in game-theoretic algorithms, modeling opponents, and detecting policy changes. In all of these settings, policy identity is the abstraction that lets us compare, retrieve, and reason about behavior \cite{grover2018learning,li2025modeling,su2022competitive,hu2024measuring,mutti2022reward,pacchiano2020learning}.

In real data, however, policy identity is usually hard to extract. Robotics play data, demonstrations, games, sports, and racing logs may provide states or state-action trajectories, but not policy labels \cite{lynch2020learning,pmlr-v229-walke23a,khazatsky2024droid,lichess,fastf1}. We study this unlabeled multi-policy setting under one assumption: the policy is fixed within an episode. No policy labels, pairwise same-policy labels, policy parameters, or environments are available during training. The representation must therefore be inferred offline and self-supervisedly, from the episode's state-action structure.

Prior work often uses stronger supervision or evaluates a different objective. Some methods assume policy labels or pairwise constraints and can train classifiers or contrastive representations \cite{chen2020simpleframeworkcontrastivelearning,ma2025contrastive}. Others learn trajectory latents with conditional variational autoencoders (CVAEs), recurrent encoders, vector-quantized (VQ) codes, diffusion models, or hypernetworks, but primarily evaluate action prediction or downstream control rather than whether the latent recovers policy identity \cite{papoudakis2021agent,he2023learning,li2025adaptively,ge2025learning,liu2020learning,lynch2020learning,co2018self,meng2023m3,kujanpaa2023hierarchical,liang2024make,hegde2024warpd,ren2025hypogen}. We instead evaluate policy representations directly: can the learned latent separate policies without labels, and does it remain useful under behavioral distribution shift?

We introduce \emph{Behavioral INR}, an implicit neural representation of behavior. Vision INRs represent an image as a function from coordinates to RGB values \cite{sitzmann2020implicit,tancik2020fourier,mildenhall2021nerf}. We represent a policy as a function from states to actions. This holds outside of pure (deterministic) policies, to which we limit our scope; in partially observed or history-dependent domains, $s$ can be understood as the available observation or compact history representation. Given an episode, a latent code modulates a shared state-action network through FiLM layers, so the latent must explain how actions vary as a function of states rather than merely summarize the trajectory \cite{perez2018film,park2019deepsdf,dupont2022data}. This also gives Behavioral INR a natural way to handle variable episode length and sampling granularity: an episode provides a set of state-action samples from an underlying function, just as an image provides coordinate-value samples from a visual signal \cite{garnelo2018neural,dupont2022data}.

We also introduce an OOD formulation for imitation-style policy datasets in general. Standard behavioral OOD settings typically vary agents, policies, tasks, or environments \cite{koh2021wilds,hendrycks2019benchmarking,sagawa2019distributionally,arjovsky2019invariant,ganin2016domain,tobin2017domain}. In multi-policy imitation data, two additional axes arise: state-distribution shift and action-distribution shift. A model can fail by identifying policies from $p(s)$ or $p(a)$ instead of the conditional map $\pi(a \mid s)$, a form of shortcut learning that is especially problematic when policies share state or action support \cite{Geirhos_2020,damour2020underspecificationpresentschallengescredibility}. Our controlled splits test this failure mode by forcing methods to recover policy identity under state/action overlap.

We evaluate on synthetic Gaussian random field data, MuJoCo demonstrations and augmented checkpoint rollouts, DM Lab Seek-Avoid, Lichess, DROID, and FastF1 \cite{minari,todorov2012mujoco,beattie2016deepmindlab,gulcehre2020rl,lichess,khazatsky2024droid,fastf1}. Behavioral INR is most effective in the harder regimes: longer episodes, more policies, and OOD splits where marginal shortcuts are weak. Across these settings, we find that Behavioral INR is strongest when policy identity must be inferred from a complex state-action function, while amortized history encoders remain competitive when labels can be recovered from state/action shortcuts, symbolic repetition, or low-dimensional action marginals.

\begin{figure}[t]
    \centering
    \includegraphics[width=\linewidth, height=4cm, keepaspectratio=false]{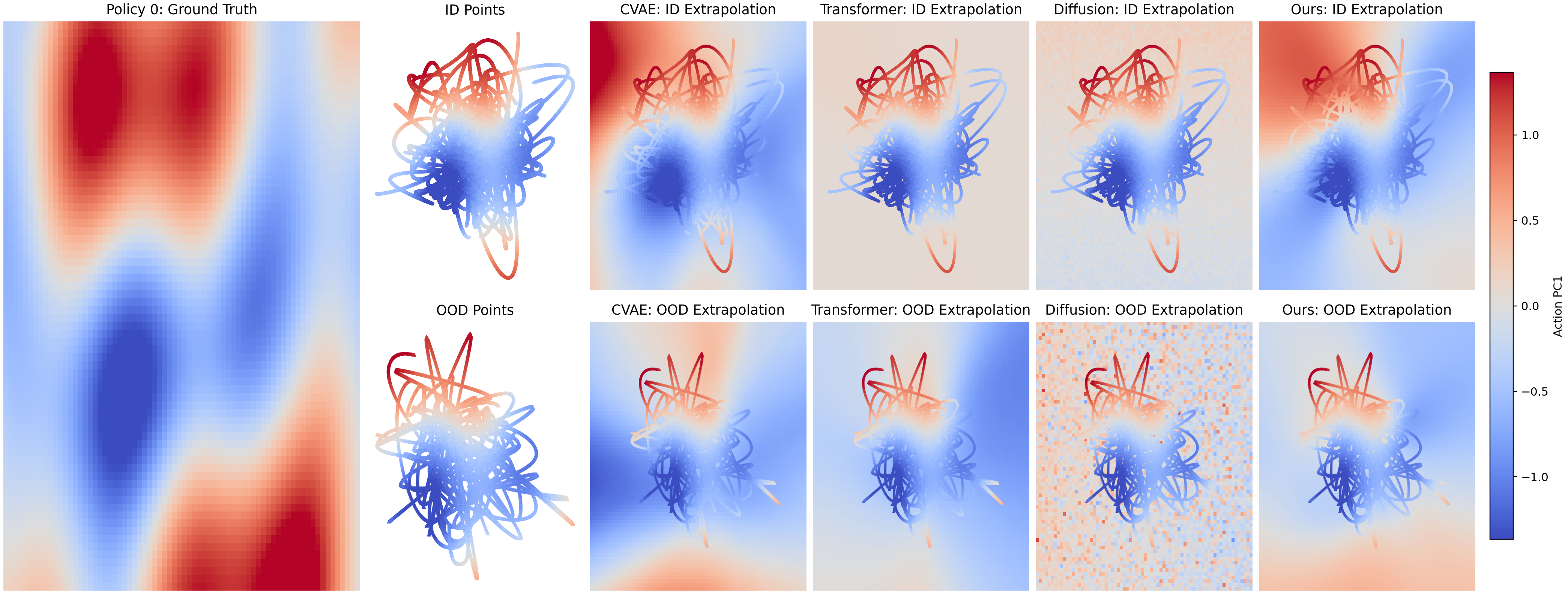}
    \caption{
    Synthetic Gaussian Random Field (GRF) data being used for in-distribution (ID) and out-of-distribution (OOD) extrapolation. Each model observes state-action pairs from an ID region and predicts actions on held-out states from the same policy. Our Behavioral INR recovers the underlying state-action function robustly.
    }
    \label{fig:grf_extrapolation}
\end{figure}

\section{Related Work}

\paragraph{Policy representations.}
Many methods use policy representations without naming them as such. Multi-agent policy representations, opponent models, partner models, theory-of-mind modules, and policy-distance methods all infer a proxy for the policy that generated behavior \cite{grover2018learning,li2025modeling,su2022competitive,wang2021tom2c,liu2020multi,ma2025contrastive,hu2024measuring,sang2022pandr,xie2021learning}. These works show that policy identity is useful for downstream interaction, but they usually evaluate control, adaptation, or prediction rather than whether a representation recovers policy identity from unlabeled episodes.

The closest methods learn trajectory or behavior latents. CAE/CVAE and recurrent variants encode state-action histories and decode actions or future trajectories \cite{papoudakis2021agent,he2023learning,li2025adaptively,ge2025learning,liu2020learning,lynch2020learning,co2018self}; VQ-style methods learn discrete trajectory codes \cite{meng2023m3,kujanpaa2023hierarchical}; and behavior-map or occupancy methods represent policies through state-action distributions \cite{pacchiano2020learning,mutti2022reward}. Hypernetwork and diffusion policy generators instead produce policy parameters from behavior prompts or task context \cite{ren2025hypogen,liang2024make,hegde2024warpd}, while policy-compression methods often assume access to checkpoint weights \cite{hegde2023generating,tenedini2025parameters,fraschini2026unsupervised}. Our setting is stricter: only unlabeled state-action episodes are available. We therefore compare to faithful adaptations of these representation mechanisms rather than methods requiring labels, rewards, or policy parameters.

\begin{figure*}[t]
\centering

\begin{minipage}[t]{0.54\textwidth}
    \centering
    \setlength{\tabcolsep}{2pt}
    
    \raisebox{-\height}{\vbox{
        \begin{tabular}{cc}
        \begin{subfigure}[t]{0.46\columnwidth}
            \centering
            \includegraphics[width=0.48\linewidth]{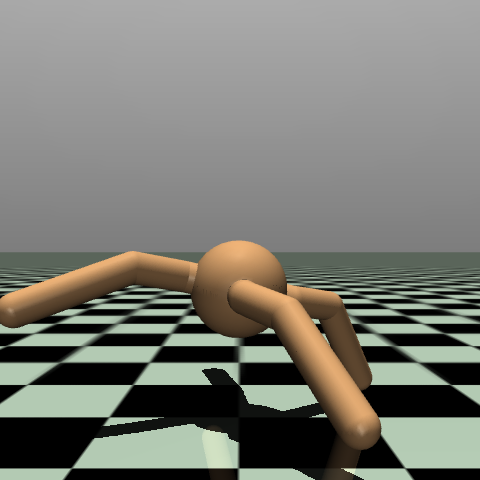}\hfill
            \includegraphics[width=0.48\linewidth]{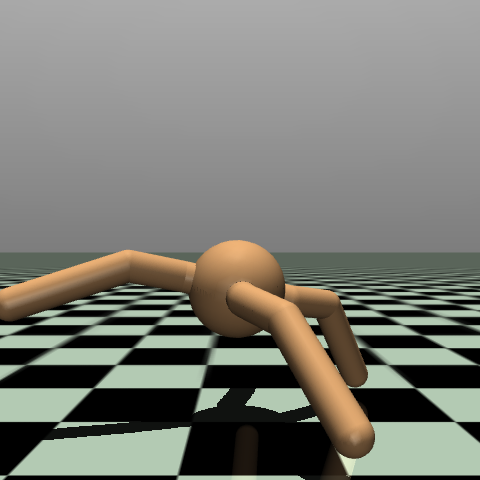}
            \caption{ID, Simple}
        \end{subfigure}
        &
        \begin{subfigure}[t]{0.46\columnwidth}
            \centering
            \includegraphics[width=0.48\linewidth]{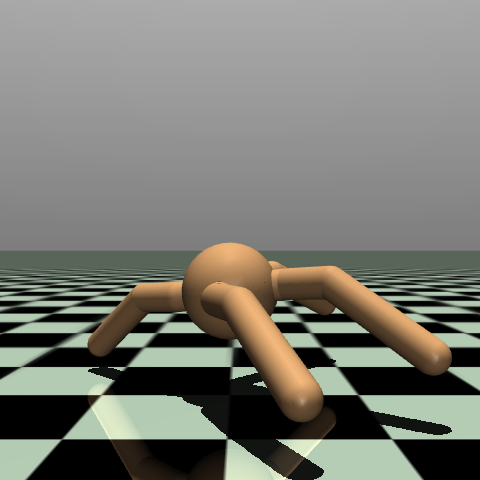}\hfill
            \includegraphics[width=0.48\linewidth]{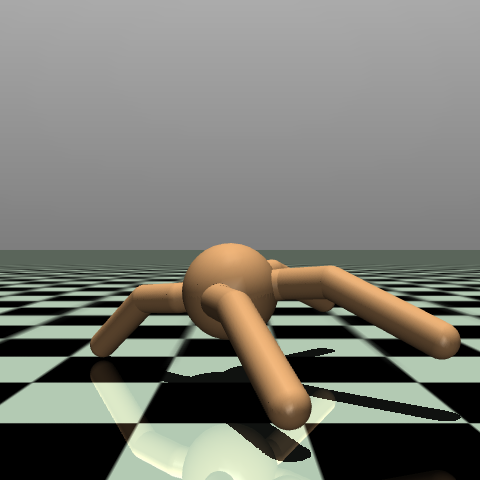}
            \caption{ID, Expert}
        \end{subfigure}
        \\[6pt]
        \begin{subfigure}[t]{0.46\columnwidth}
            \centering
            \includegraphics[width=0.48\linewidth]{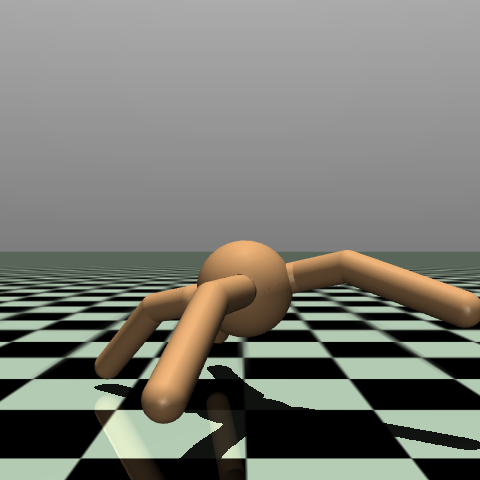}\hfill
            \includegraphics[width=0.48\linewidth]{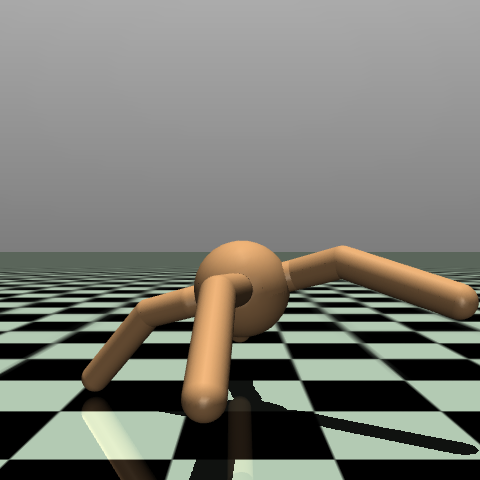}
            \caption{OOD, Simple}
        \end{subfigure}
        &
        \begin{subfigure}[t]{0.46\columnwidth}
            \centering
            \includegraphics[width=0.48\linewidth]{figs/mujocoAntOOD/ant/ssid0000_t0064_state.png}\hfill
            \includegraphics[width=0.48\linewidth]{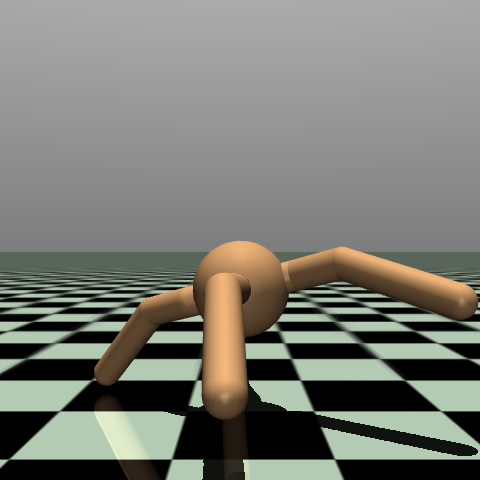}
            \caption{OOD, Expert}
        \end{subfigure}
        \end{tabular}
    }}

    \vspace{10pt} 
    \caption{
    Ant sequences from Minari \cite{minari}. We construct out-of-distribution (OOD) sequences by sampling state-action pairs based on action similarity across policies, such as simple, medium, and expert. Each panel shows two frames from a sequence. In the OOD split, action-similar samples can correspond to visually distinct or unrecoverable Ant states, making policy identity difficult to infer from state marginals alone.
    }
    \label{fig:mujoco_ant}
\end{minipage}%
\hfill 
\begin{minipage}[t]{0.42\textwidth}
    \centering
    
    \raisebox{-\height}{\vbox{
        \includegraphics[width=0.95\linewidth]{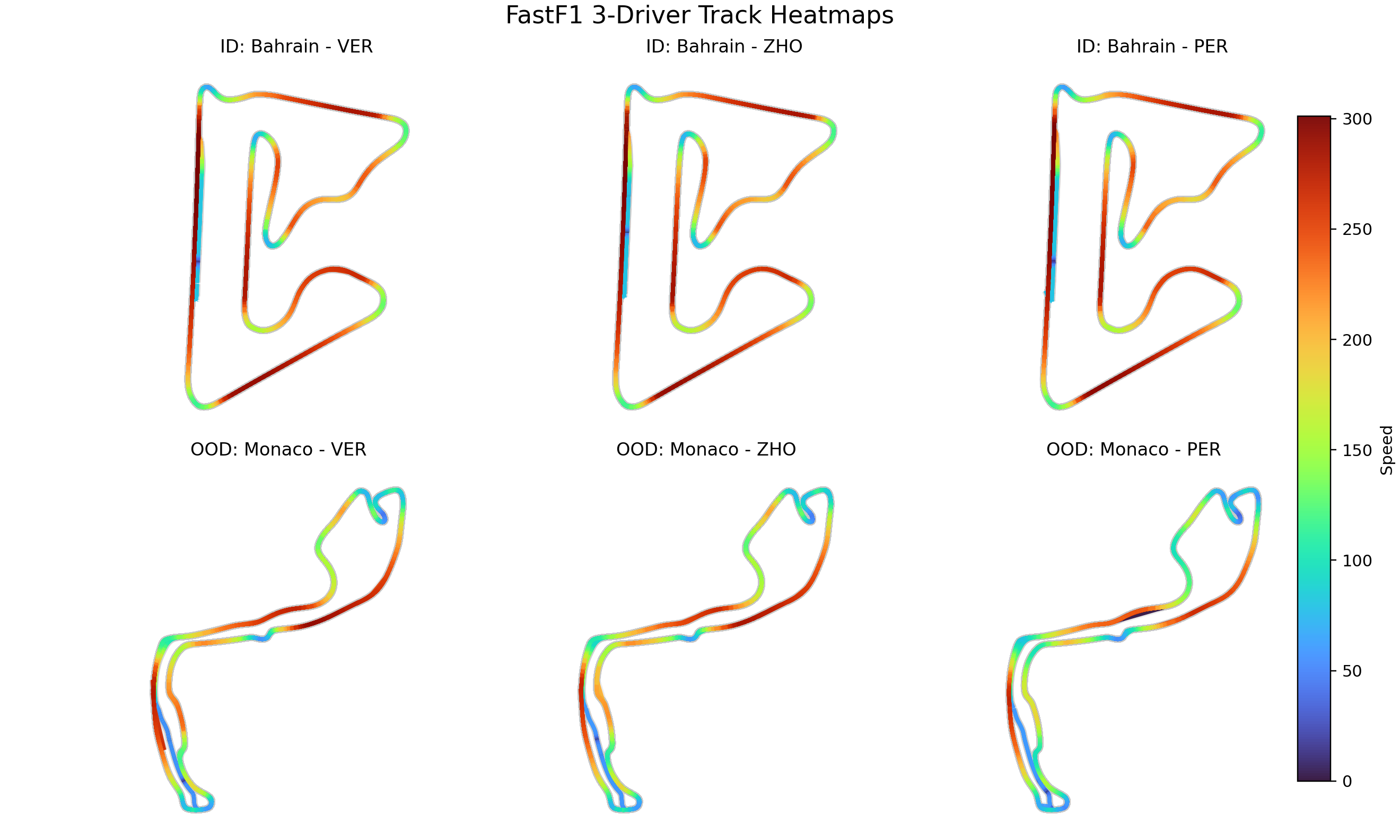}
    }}
    
    \vspace{10pt}
    \caption{
    We use Formula One racing data (telemetry) as one of our real-world datasets. We set the Monaco Grand Prix as the OOD split, and other tracks (including Bahrain) as the ID split. In this figure, we show the ID/OOD heatmaps of Max Verstappen, Zhou Guanyu, and Sergio Perez.
    }
    \label{fig:f1_heatmap}
\end{minipage}

\end{figure*}

\paragraph{Imitation and heterogeneous demonstrations.}
Behavioral cloning and imitation learning model demonstrations for action prediction or control. Prior work addresses covariate shift through data aggregation and perturbations \cite{ross2011reduction,laskey2017dart}, occupancy matching \cite{ho2016generative}, and learning from imperfect, imbalanced, or shifted demonstrations \cite{wu2019imitation,bashiri2021distributionally,xu2022discriminator,fu2023ess,parekh2025towards}. Large robotics datasets and offline RL benchmarks emphasize scale, task diversity, and support constraints \cite{pmlr-v229-walke23a,9788026,10611477,pmlr-v164-jang22a,liu2023libero,fujimoto2019off,zhou2021plas}. These works ask how to learn a good policy from data; we ask whether unlabeled data contains recoverable policy identities and how robust those identities are under behavioral shift.

\paragraph{Implicit neural representations and generative clustering.}
INRs model data as continuous functions, such as coordinate-to-signal maps in images, shapes, and neural fields \cite{sitzmann2020implicit,tancik2020fourier,mildenhall2021nerf}. Conditioning methods such as hypernetworks and FiLM modulate shared function priors \cite{ha2016hypernetworks,perez2018film}, while neural processes and Functa-style models treat datapoints as function instances \cite{garnelo2018neural,dupont2022data}. Behavioral INR applies this view to policies: an episode is a sample from a state-action function. Our self-supervised objective is also related to generative classification and clustering, where labels are inferred from latent variables that explain the data \cite{ng2001discriminative,kingma2013auto,dilokthanakul2016deep,jiang2016variational}.

\begin{figure}[t]
    \centering
    \setlength{\fboxrule}{1.5pt}
    \setlength{\fboxsep}{0pt}

    \hbox to \columnwidth{%
        \begin{subfigure}[t]{0.23\columnwidth}
            \centering
            \vspace{0pt}
            \includegraphics[width=\linewidth]{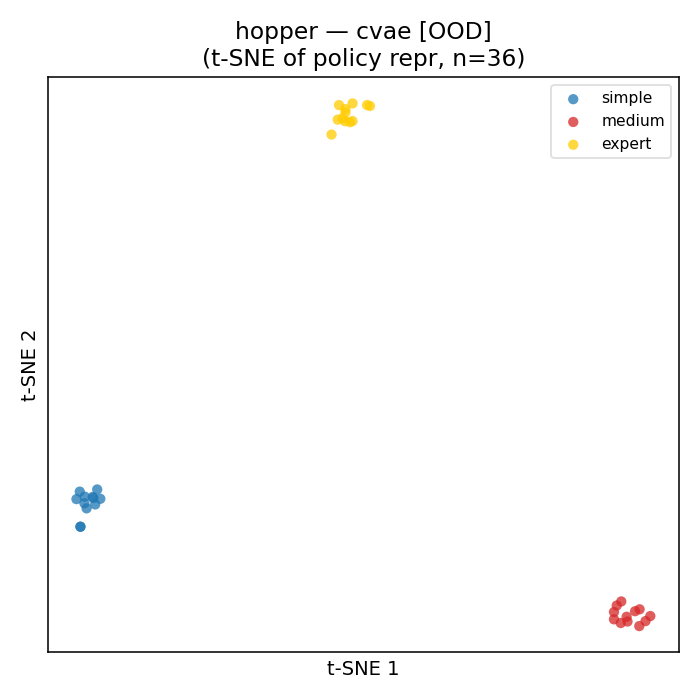}\par
            \vspace{2pt}
            \includegraphics[width=\linewidth]{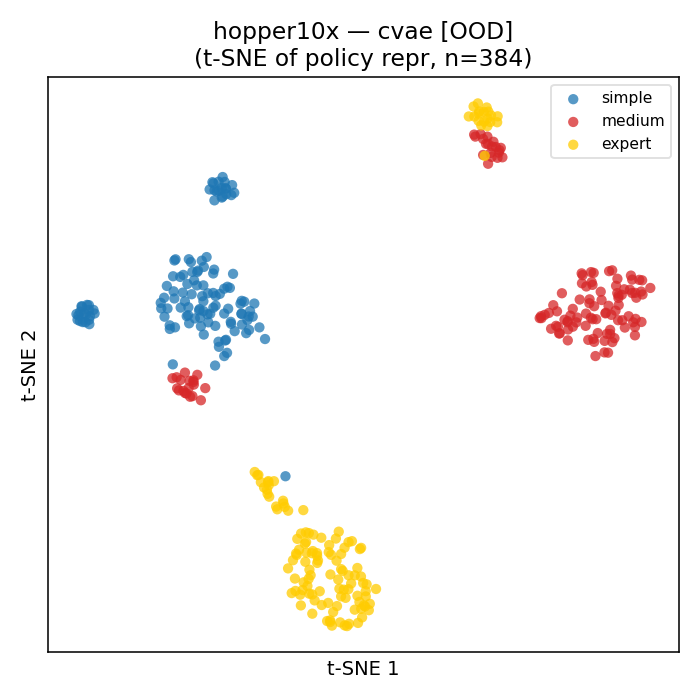}\par
            \vspace{2pt}
            \includegraphics[width=\linewidth]{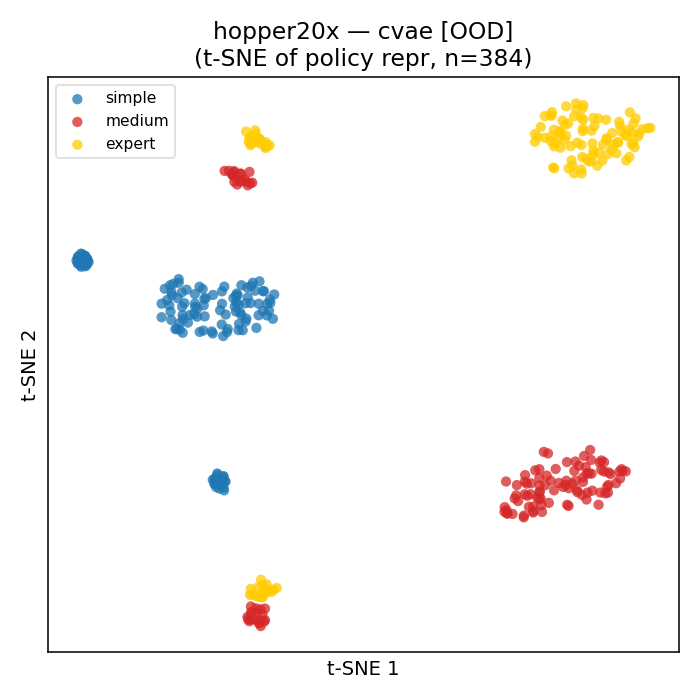}
            \caption{CVAE.}
        \end{subfigure}%
        \hfill
        \begin{subfigure}[t]{0.23\columnwidth}
            \centering
            \vspace{0pt}
            \includegraphics[width=\linewidth]{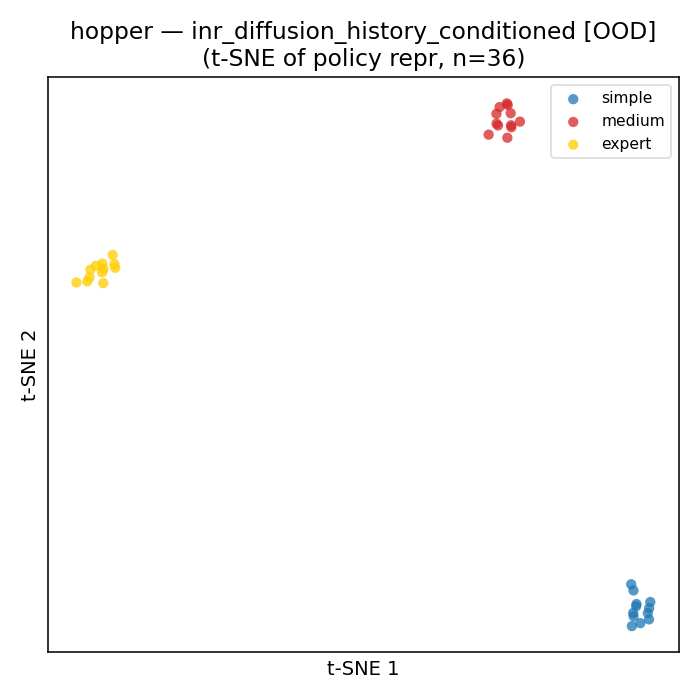}\par
            \vspace{2pt}
            \includegraphics[width=\linewidth]{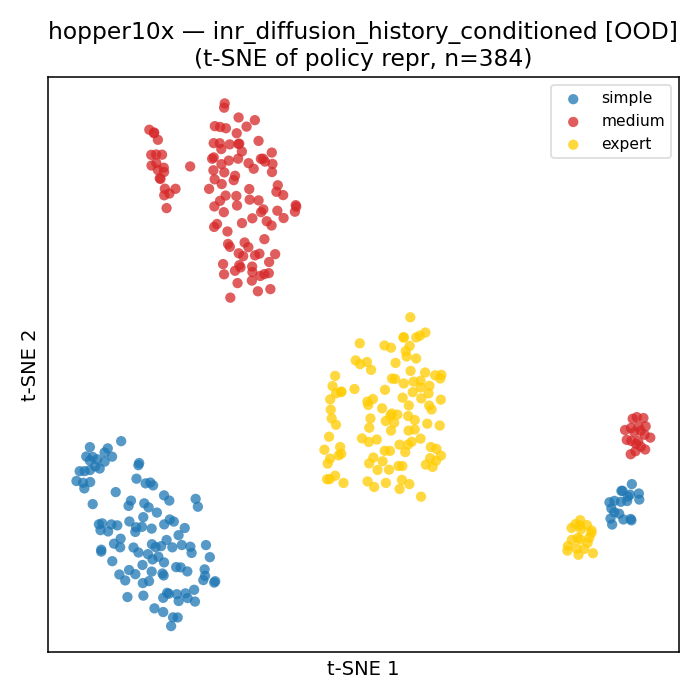}\par
            \vspace{2pt}
            \includegraphics[width=\linewidth]{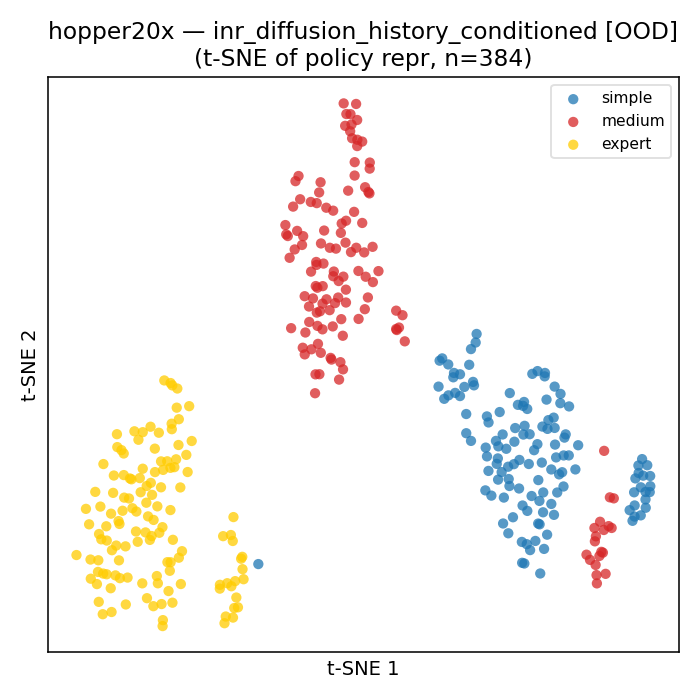}
            \caption{Diff.}
        \end{subfigure}%
        \hfill
        \begin{subfigure}[t]{0.23\columnwidth}
            \centering
            \vspace{0pt}
            \includegraphics[width=\linewidth]{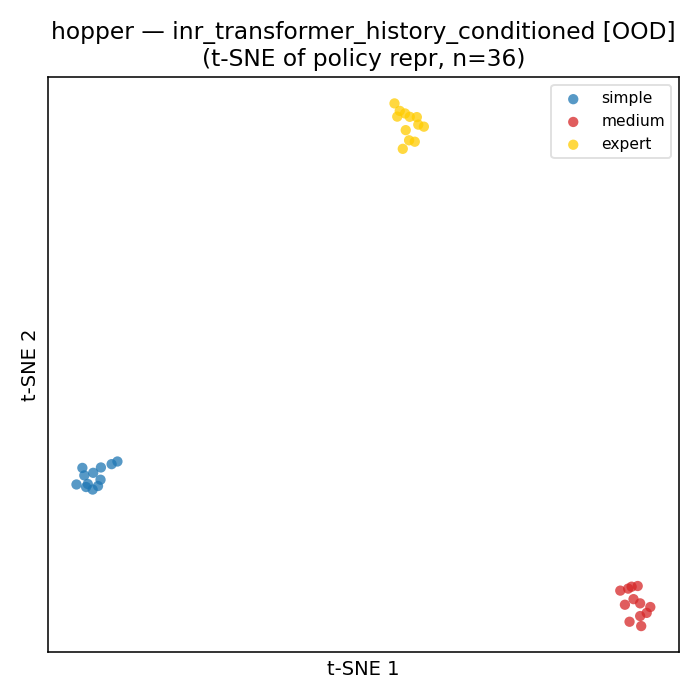}\par
            \vspace{2pt}
            \includegraphics[width=\linewidth]{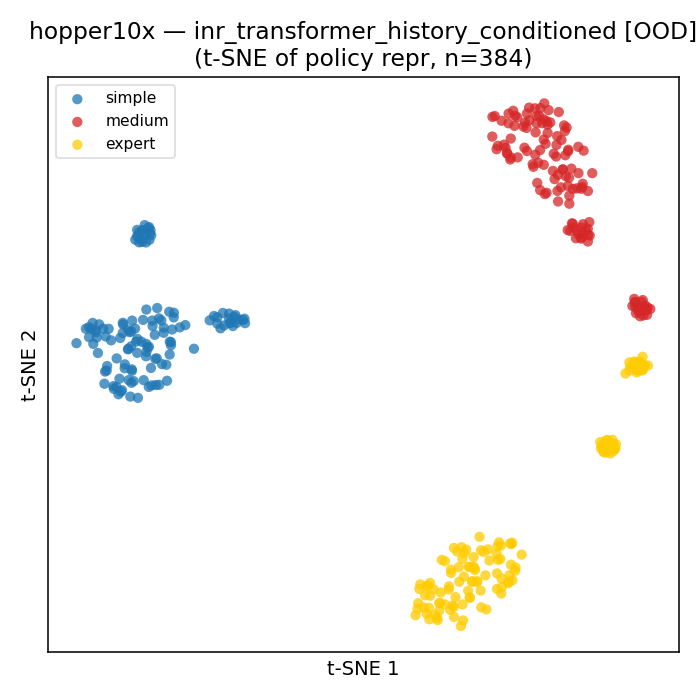}\par
            \vspace{2pt}
            \includegraphics[width=\linewidth]{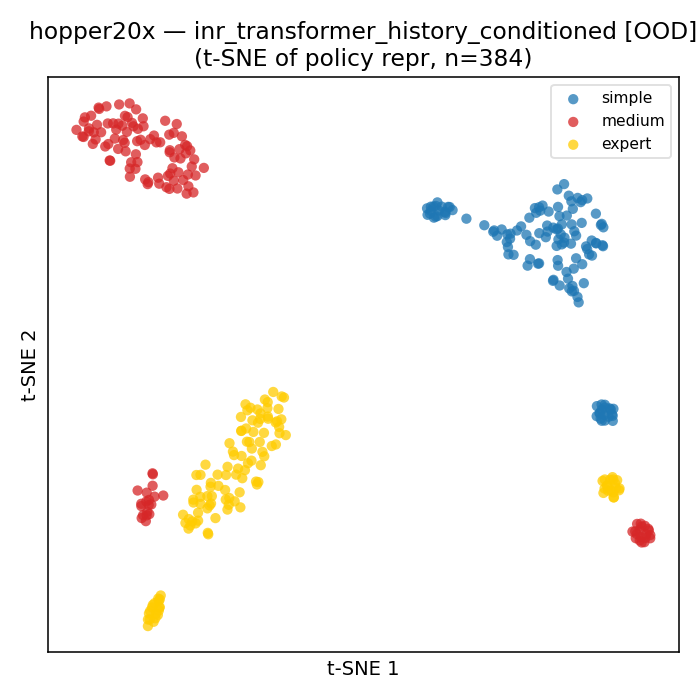}
            \caption{Transf.}
        \end{subfigure}%
        \hfill
        \begin{subfigure}[t]{0.23\columnwidth} 
            \centering
            \vspace{0pt}
            \fbox{%
                \begin{minipage}{\dimexpr\linewidth-2\fboxrule-2\fboxsep\relax}
                    \centering
                    \includegraphics[width=\linewidth]{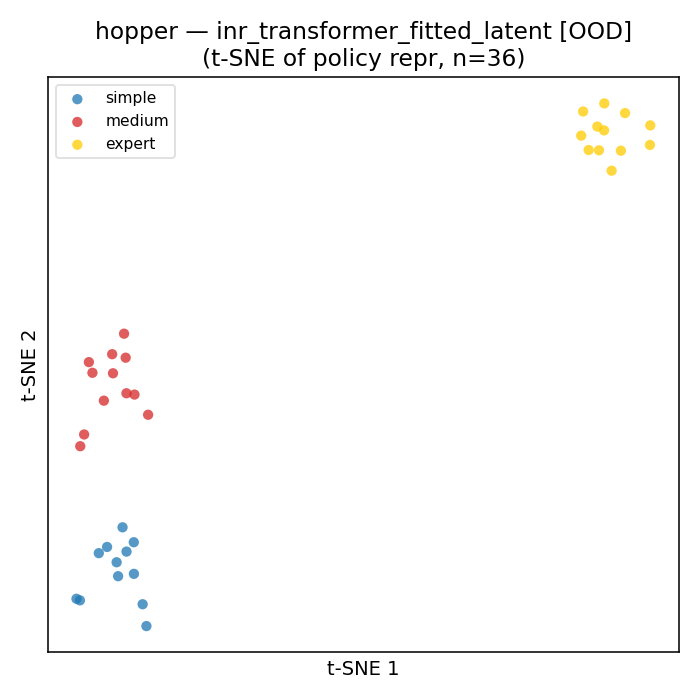}\par
                    \vspace{2.5pt}
                    \includegraphics[width=\linewidth]{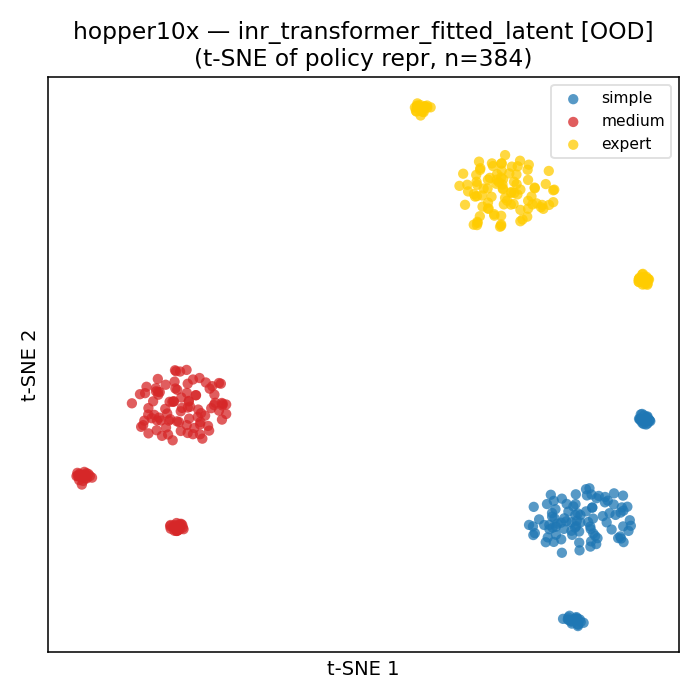}\par
                    \vspace{2.5pt}
                    \includegraphics[width=\linewidth]{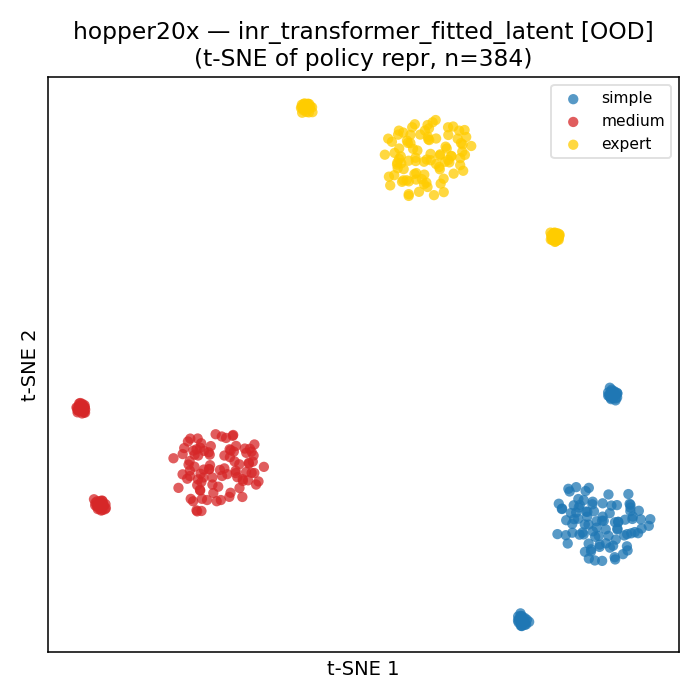}
                \end{minipage}%
            }%
            \par\vspace{2pt}
            \caption{Ours}
        \end{subfigure}%
    }

    \caption{Out-of-distribution policy representations on Hopper at increasing data scales. Behavioral INR remains visually separable at larger Hopper scales, while amortized history-conditioned representations degrade, matching the probe-accuracy trend in Table~\ref{tab:hopper_scaling_extended}.}
    \label{fig:hopper_clusters}
\end{figure}

\section{Approach}

\paragraph{Problem.}
A dataset consists of episodes
\[
\tau_i=\{(s_{i,t},a_{i,t}\}_{t=1}^{T_i},
\]
where each episode is generated by an unknown fixed policy $\pi_i$. During training, the model observes state-action episodes and episode boundaries, but not policy labels. It learns a representation $z_i$ that should both predict actions and recover policy identity. Policy labels are used only after training for linear-probe and $k$NN evaluation.

\paragraph{Policy-level OOD.}
We evaluate whether $z_i$ captures $\pi(a\mid s)$ rather than shortcuts in $p(s)$ or $p(a)$. We therefore define policy-level OOD along two axes: state-distribution shift, where the same policy is evaluated on different states, and action-distribution shift, where marginal action statistics change across train/test. Unlike standard OOD evaluation that changes the environment or policy set, our splits hold policy identity fixed and shift the observed support of states or actions, testing whether a representation identifies the conditional map rather than marginal visitation statistics. These shifts are most diagnostic when policies overlap in states or actions. For synthetic and checkpoint-generated data, ID/OOD partitions are constructed directly from the known generator. For real-world datasets, we use domain-specific OOD partitions and treat a player, driver, or demonstrator as the policy proxy.

\paragraph{Data.}
We evaluate on Synthetic GRF, MuJoCo/Minari, augmented MuJoCo Hopper, DM Lab Seek-Avoid, Lichess, DROID, and FastF1. Synthetic GRF gives a controlled nonlinear state-action function. MuJoCo uses fixed policy checkpoints, allowing controlled state/action splits and 1x/10x/20x episode-scale variants. DMLab tests discrete-action visual behavior. Lichess, DROID, and FastF1 test real-world domains with player, demonstrator, and driver identities, respectively. Unless otherwise stated, all main experiments use a 64-dimensional representation.

\begin{table*}[t]
\centering
\small
\caption{
Baseline families and how they are instantiated in our setting. Prior work often assumes additional inputs, such as policy labels, pairwise labels, checkpoint parameters, or downstream rewards. We compare to faithful adaptations that use only the information available in our problem: unlabeled state-action episodes.
}
\label{tab:baseline_taxonomy}
\setlength{\tabcolsep}{5pt}
\begin{tabular}{p{0.20\textwidth}p{0.34\textwidth}p{0.37\textwidth}}
\toprule
Family & Representative prior work & Baseline in this paper \\
\midrule
CAE/CVAE trajectory latent &
Agent modeling, assistive imitation, latent strategies, trajectory embeddings, ice hockey, latent plans, SeCTAR
\cite{papoudakis2021agent,he2023learning,li2025adaptively,ge2025learning,liu2020learning,lynch2020learning,co2018self} &
CVAE-Transf. and CVAE-RNN encode state-action history and decode action conditioned on current state. \\
\midrule
Discrete trajectory code &
VQ/modular multi-agent pretraining and hierarchical imitation
\cite{meng2023m3,kujanpaa2023hierarchical} &
VQ-VAE encodes each episode into a discrete latent code before action decoding. \\
\midrule
Distributional behavior embedding &
Behavior embedding maps and occupancy-style policy compression
\cite{pacchiano2020learning,mutti2022reward} &
State-action distribution baseline using marginal or occupancy-like behavior statistics. \\
\midrule
History-conditioned policy generation &
HypoGen, Make-An-Agent, WARPD, and related policy generators
\cite{ren2025hypogen,liang2024make,hegde2024warpd} &
History-conditioned INR Transformer and history-conditioned INR diffusion. These adapt the conditioning mechanism without requiring policy checkpoint weights. \\
\midrule
Checkpoint policy compression &
Latent diffusion or low-dimensional compression of policy parameters
\cite{hegde2023generating,tenedini2025parameters,fraschini2026unsupervised} &
Not directly comparable: these methods require access to policy network parameters, which are unavailable in our unlabeled episode setting. \\
\midrule
Fitted functional latent &
DeepSDF/Functa-style fitted representations
\cite{park2019deepsdf,dupont2022data} &
Behavioral INR fits an episode-level latent that modulates a shared state-to-action function. \\
\bottomrule
\end{tabular}
\end{table*}

\paragraph{Architectures and training.}
As described in the Appendix, all four main architectures are trained with the same offline batches of past state-action history, current state, and next action. Continuous-action domains use regression loss; discrete-action domains use cross-entropy. We train with AdamW for 30 epochs, batch size 256, evaluation batch size 512, learning rate $3\cdot10^{-4}$, weight decay $10^{-4}$, gradient clipping at 1.0, mixed precision, and no policy labels. The default representation dimension is 64, with $d_{\mathrm{model}}=128$, 4 attention heads, and dropout 0.

\emph{CVAE (CVAE-Transf.).}
The CVAE-Transf. uses a permutation-invariant Transformer encoder over past state-action pairs. The encoder outputs $(\mu,\log\sigma^2)$, and the representation used for evaluation is $\mu$. A decoder receives the sampled latent and an embedding of the current state, then predicts the next action. The objective is reconstruction loss plus $10^{-2}$ times the KL term.

\emph{History-conditioned INR Transformer (INR Transf.).}
This baseline amortizes the policy latent from ordered past state-action history. A Transformer encoder maps the history to $z$, and a FiLM-modulated MLP maps the current state to the next action. It tests whether an INR-style decoder helps when the latent is still produced by a standard history encoder.

\emph{History-conditioned INR Diffusion (INR Diff.).}
This model uses the same history-conditioned factorization, but predicts actions with a conditional DDPM-style epsilon predictor. We use 50 diffusion steps with a linear beta schedule from $10^{-4}$ to $0.02$ and 10 sampling steps at evaluation. This baseline adapts policy-generation ideas to our episode-conditioned action-prediction setting.

\emph{Behavioral INR (Ours).}
Behavioral INR assigns a learnable latent to each training behavior unit and uses that latent to FiLM-modulate a shared state-to-action INR. The decoder is a 3-block FiLM MLP with hidden size 256. At test time, shared INR weights are frozen and only the episode latent is optimized on the support state-action pairs. We use 40 latent-inference steps with learning rate $5\cdot10^{-2}$ and $10^{-4}$ latent $\ell_2$ regularization. Thus, representation extraction is function fitting: $z_i$ is the code that best explains the observed state-action map.

\emph{Additional Baselines.}
Additional baselines include a recurrent CVAE encoder (CVAE-RNN), a VQ-VAE trajectory-code model, and a state-action behavior embedding map (BEM). All three use the same train/test splits and evaluation protocol; BEM is non-generative, so action-prediction losses are not reported. CVAE-RNN replaces the permutation-invariant Transformer encoder of the CVAE with a 2-layer unidirectional GRU of hidden size $128$; the final-layer hidden state is projected to $\mathbb{R}^{64}$. The VQ-VAE shares the CVAE-Transf.'s encoder but replaces the Gaussian bottleneck with vector quantization: the encoder produces a $64$-dim continuous latent that is split into $4$ slots, each independently quantized against a shared codebook of $256$ entries with slot dimension $16$; training combines the action reconstruction loss with the standard codebook plus commitment objective ($\beta=0.25$) under a straight-through gradient estimator. BEM is non-parametric: we fit $k$-means with $k{=}64$ once on the training $(s,a)$ pairs ($50$ iterations, seeded), and given an episode window of past history plus the current state-action, we return the $L_1$-normalized count vector over the $64$ clusters as the representation; this window-level histogram is averaged across the per-episode samples.

\paragraph{Evaluation.}
We report linear probe accuracy and $k$NN accuracy for policy identity recovery, using labels only after training. We also report generative metrics: NMSE and median squared error for continuous actions, and action accuracy/NLL for discrete actions. Median squared error is reported because rare large predictions can inflate mean error. We sweep seven ID/OOD configurations: \textsc{no-shift}, \textsc{new-policy}, \textsc{single-shift}, \textsc{conflation}, \textsc{generalization}, \textsc{specialization}, and \textsc{novel-generalization}, as summarized in Table~\ref{tab:compExperiments}.


\section{Results}

We evaluate two properties: whether the representation recovers policy identity, and whether it predicts actions under OOD behavioral shift. Since no standard benchmark exists for unlabeled policy identity recovery, our baselines instantiate the main comparable mechanisms: latent trajectory autoencoding, history-conditioned INR decoding, and history-conditioned diffusion. All methods are trained without policy labels; labels are used only for linear probes and $k$NN evaluation. 

We therefore organize the results by how much each setting requires recovery of the conditional state-action function, rather than by treating every dataset as an equally diagnostic benchmark. Settings with separable state marginals, repeated symbolic contexts, or low-dimensional action shortcuts can make amortized history encoders competitive; settings with longer episodes, more policies, and weaker shortcuts should favor fitted functional representations.

\paragraph{Synthetic GRF.}
Synthetic GRF policies provide controlled nonlinear state-action functions. Under \textsc{specialization}, train episodes cover ID states and test episodes evaluate the same policies on OOD states. Table~\ref{tab:grf_generalization_results} shows that Behavioral INR gives the highest linear probe accuracy, indicating that fitting a state-action function makes policy identity more linearly recoverable under state shift. The GRF visualization in Figure~\ref{fig:grf_extrapolation} shows the same mechanism qualitatively: Behavioral INR extrapolates the policy function more coherently outside the observed region.

\begin{table}[t]
\centering
\small
\caption{
Synthetic GRF 10x all-policy results under the \textsc{generalization} split. Values are mean $\pm$ standard deviation over seeds. Higher is better.
}
\label{tab:grf_generalization_results}
\setlength{\tabcolsep}{3pt}

\resizebox{\columnwidth}{!}{%
  \begin{tabular}{lccc}
  \toprule
  Model & Probe Acc. $\uparrow$ & $k$NN-5 $\uparrow$ & $k$NN-1 $\uparrow$ \\
  \midrule
  CVAE-Transf. & $0.567 \pm 0.141$ & $0.867 \pm 0.031$ & $0.433 \pm 0.079$ \\
  CVAE-RNN & $0.533 \pm 0.094$ & $0.800 \pm 0.094$ & $0.378 \pm 0.094$ \\
  VQ-VAE & $\mathbf{0.611 \pm 0.047}$ & $\mathbf{0.878 \pm 0.016}$ & $\mathbf{0.589 \pm 0.047}$ \\
  State-action BEM & $0.489 \pm 0.220$ & $0.844 \pm 0.031$ & $0.389 \pm 0.141$ \\
  \bottomrule
  \end{tabular}%
  \qquad 
  \begin{tabular}{lccc}
  \toprule
  Model & Probe Acc. $\uparrow$ & $k$NN-5 $\uparrow$ & $k$NN-1 $\uparrow$ \\
  \midrule
  INR-Transf. & $0.500 \pm 0.016$ & $0.789 \pm 0.047$ & $0.333 \pm 0.031$ \\
  INR-Diff. & $0.544 \pm 0.016$ & $0.800 \pm 0.000$ & $0.433 \pm 0.236$ \\
  \textbf{Ours} & $\mathbf{0.611 \pm 0.110}$ & $0.778 \pm 0.031$ & $0.478 \pm 0.173$ \\
  \\ 
  \bottomrule
  \end{tabular}%
}

\end{table}
\paragraph{MuJoCo scaling.}
Public Minari splits are less diagnostic because simple, medium, and expert checkpoints often occupy separable state regions; several methods can recover policy identity from these state-distribution cues. We therefore use augmented Hopper suites with longer checkpoint rollouts and harder action-resampled splits. Table~\ref{tab:hopper_scaling_extended} shows the resulting scaling pattern: Behavioral INR preserves perfect probe accuracy at 10x and 20x, while amortized baselines degrade. Pointwise action loss and policy identifiability measure different properties: a history-conditioned predictor can obtain low action error from local correlations or marginal shortcuts while producing a latent that does not separate policies under OOD specialization. The key result is that fitted latents preserve policy identity as episode length and OOD difficulty increase.

\paragraph{DMLab and real-world domains.}
DMLab Seek-Avoid is a discrete-action sanity check; representation metrics are nearly saturated, so we do not treat it as the main scaling evidence. For real-world datasets, policy identity is proxied by player, demonstrator, or driver. Lichess uses held-out game contexts, DROID uses held-out manipulation task families, and FastF1 uses held-out racing contexts. Table~\ref{tab:real_world_specialization} shows that Behavioral INR is strongest on FastF1, ties or competes on DROID, and is weaker on Lichess. FastF1 is the clearest real-world setting where Behavioral INR consistently ranks first, although absolute accuracies remain low because all-driver classification is substantially harder than controlled checkpoint classification.





\begin{table*}[t]
\centering
\scriptsize

\begin{minipage}[t]{0.48\textwidth}
\centering
\setlength{\tabcolsep}{3pt}
\caption{
Hopper scaling (starting from public offline data \cite{minari} under the \textsc{specialization} split. Values are mean $\pm$ standard deviation over two seeds from the aggregate table. Higher is better for probe accuracy; lower is better for median squared error (MedSE). As the episode length and count increase, we observe that Behavioral INR (Ours) dominates probe accuracy, while being on par in action prediction.
}
\label{tab:hopper_scaling_extended}
\begin{tabular}{llccc}
\toprule
Setting & Model & Probe $\uparrow$ & MedSE $\downarrow$ \\
\midrule
Hopper
& CVAE-Transf.
& $\mathbf{1.000 \pm 0.000}$
& $0.320 \pm 0.003$ \\
& INR-Transf.
& $\mathbf{1.000 \pm 0.000}$
& $\mathbf{0.173 \pm 0.006}$ \\
& INR-Diff.
& $\mathbf{1.000 \pm 0.000}$
& $0.929 \pm 0.063$ \\
& \textbf{Ours}
& $\mathbf{1.000 \pm 0.000}$
& $0.179 \pm 0.008$ \\
\midrule
Hopper 10x
& CVAE-Transf.
& $\mathbf{1.000 \pm 0.000}$
& $0.259 \pm 0.001$ \\
& INR-Transf.
& $\mathbf{1.000 \pm 0.000}$
& $\mathbf{0.182 \pm 0.005}$ \\
& INR-Diff.
& $0.750 \pm 0.354$
& $0.990 \pm 0.016$ \\
& \textbf{Ours}
& $\mathbf{1.000 \pm 0.000}$
& $0.244 \pm 0.003$ \\
\midrule
Hopper 20x
& CVAE-Transf.
& $0.763 \pm 0.223$
& $\mathbf{0.313 \pm 0.013}$ \\
& INR-Transf.
& $0.618 \pm 0.168$
& $0.387 \pm 0.044$ \\
& INR-Diff.
& $0.500 \pm 0.000$
& $0.995 \pm 0.030$ \\
& \textbf{Ours}
& $\mathbf{1.000 \pm 0.000}$
& $0.398 \pm 0.123$ \\
\bottomrule
\end{tabular}
\end{minipage}
\hfill 
\begin{minipage}[t]{0.48\textwidth}
\centering
\setlength{\tabcolsep}{4pt}
\caption{
All-policy \textsc{specialization} results on real-world domains. Values are mean probe and $k$NN-1 accuracy over seeds. These datasets differ in shortcut structure: FastF1 is the clearest continuous state-action scaling setting, DROID uses compact low-dimensional features where action shortcuts are informative, and Lichess is a symbolic discrete-action domain with repeated contexts.
}
\label{tab:real_world_specialization}
\begin{tabular}{llcc}
\toprule
Domain & Model  & Probe Acc. $\uparrow$ & $k$NN-1 Acc. $\uparrow$ \\
\midrule
FastF1 & CVAE-Transf.  & 0.143 & 0.071 \\
FastF1 & CVAE-RNN & 0.151 & 0.067\\
FastF1 & VQ-VAE & 0.106 & 0.052 \\
FastF1 & State-action BEM & 0.119 & 0.060 \\
FastF1 & INR-Transf. & 0.119 & 0.048 \\
FastF1 & INR-Diff. & 0.071 & 0.024 \\
FastF1 & \textbf{Ours}  & \textbf{0.190} & \textbf{0.119} \\
\midrule
DROID & CVAE-Transf. &  0.750 & 0.500 \\
DROID & CVAE-RNN & 0.622 & 0.500 \\
DROID & VQ-VAE &  0.500 & 0.667 \\
DROID & State-action BEM  & 0.622 & 0.667 \\
DROID & INR-Transf. & 0.500 & \textbf{0.750} \\
DROID & INR-Diff.  & \textbf{1.000} & \textbf{0.750} \\
DROID & \textbf{Ours} &  0.500 & \textbf{0.750} \\
\midrule
Lichess & CVAE-Transf.  & \textbf{0.6957} & \textbf{0.7295} \\
Lichess & CVAE-RNN  & 0.6649 & 0.6822 \\
Lichess & VQ-VAE  & 0.4931 & 0.4530 \\
Lichess & State-action BEM & 0.4858 & 0.5061 \\
Lichess & INR-Transf. & 0.6763 & 0.5652 \\
Lichess & INR-Diff. &  0.5797 & 0.4928 \\
Lichess & \textbf{Ours} & 0.5024 & 0.5072 \\
\bottomrule
\end{tabular}
\end{minipage}

\end{table*}

\paragraph{Shortcut structure explains domain-dependent probe accuracy.}
The real-world and discrete-control results should be interpreted by the shortcut structure of each domain. DM Lab Seek-Avoid is comparatively easy for policy identification: most methods reach near-saturated probe and $k$NN accuracy, so it is useful as a cross-domain sanity check but not as the strongest evidence for scalability. Lichess is difficult for a different reason: it has a large discrete action space, symbolic board states, repeated openings, and many near-equivalent tactical positions, making policy identity highly entangled with sparse move distributions and repeated contexts. DROID is also not a pure visual state-action generalization test in our setup because we use compact low-dimensional features rather than raw visual observations; action marginals and demonstrator-specific action shortcuts can therefore be highly informative. These cases clarify the main empirical pattern: Behavioral INR is strongest when the task requires recovering a complex state-conditioned action function, while amortized history encoders remain competitive when policy identity is recoverable from shortcuts.

\paragraph{Takeaway.}
Overall, the results do not show that Behavioral INR dominates in every domain. They show a sharper pattern: Behavioral INR is best when policy identity must be inferred from a complex state-action function, while amortized history encoders remain competitive when labels can be recovered from state/action shortcuts, symbolic repetition, or low-dimensional action marginals.


\section{Limitations and Future Work}

First, Behavioral INR explores one point in a larger design space. We use a Transformer/MLP INR with FiLM modulation, but future work should study hypernetwork-generated weights, diffusion priors over policy latents, hierarchical modulation, mixture-of-expert INRs, and amortized-fitted hybrids. Second, our experiments use mostly compact state representations. Extending Behavioral INR to raw visual observations, language-conditioned observations, or multimodal streams would test whether policy identity can be recovered jointly with perception. Multi-agent settings raise a similar issue: a cooperative team policy may map joint observations and communication histories to joint actions, requiring structured multi-agent INRs rather than a single-agent state-action function. Third, we assume one fixed policy per episode. Real behavior may switch between strategies or compose multiple skills. Future work could model episodes as mixtures of sub-policies with segment-level latents, connecting this formulation to change-point detection, option discovery, and hierarchical imitation learning. Another extension is uncertainty-aware modeling: some states are more diagnostic of policy identity than others, and some policies are more stochastic in specific regions of state space. Fourth, we evaluate representations through clustering and OOD action prediction. Downstream applications remain to be tested, including opponent modeling, policy-space search, data filtering, matchup prediction, and policy-space response oracles. Beyond behavior, the same fitted-function perspective may apply to scientific domains where each observed system is generated by an unknown latent mechanism, connecting neural INRs to system identification and symbolic regression.


\clearpage
\acknowledgments{This material is based upon work supported by the National Science Foundation Graduate Research Fellowship Program under Grant No. DGE2140739. Any opinions,
findings, and conclusions or recommendations expressed in this material are those of the authors and do not necessarily reflect the views of the National Science Foundation.

We also thank the organizers and participants of the ICML 2026 Workshop on Structured Probabilistic Inference \& Generative Modeling (SPIGM) for helpful feedback on an earlier version of this work.}


\bibliography{example}  

\newpage

\appendix
\section{Appendix}
\paragraph{Implementation Details.}

All models use the same data interface: a sample contains past states, past actions, current state, next action, episode/unit ID, policy ID, and an OOD flag. Policy ID is excluded from training and used only for evaluation. Continuous-action datasets use regression losses and report NMSE/MedSE; discrete-action datasets use cross-entropy and report accuracy/NLL. We use history length $K=16$, latent dimension 64, batch size 256, evaluation batch size 512, AdamW with learning rate $3\cdot10^{-4}$, weight decay $10^{-4}$, gradient clipping at 1.0, mixed precision, and 30 epochs unless otherwise specified.

For CVAE, the past state-action history is shuffled during training and encoded as a bag of pairs. For INR-based models, we also find empirically that shuffled history is beneficial. The history-conditioned Transformer and diffusion models infer $z$ amortized from history. Behavioral INR instead learns or infers $z$ by optimizing the episode latent while keeping the shared INR fixed at test time. For fitted-latent inference we use 40 steps, learning rate $5\cdot10^{-2}$, and latent $\ell_2$ penalty $10^{-4}$.

We train all models offline with the same train/validation/test partitions and never use policy labels during training. Hyperparameters, dataset caches, launch scripts, checkpoints, per-run configs, metrics logs, summaries, and aggregates are released with the code. Each run stores its composed Hydra config, checkpoint, metrics log, evaluation file, and summary. We report seed-averaged metrics when multiple seeds are available.

\paragraph{Assets and Licenses.}

We use public benchmark or open datasets: Minari/MuJoCo, RL Unplugged/DeepMind Lab, DROID, Lichess, and FastF1. We cite each dataset or software package in the main paper. The released code will include scripts for downloading or rebuilding caches when redistribution is not appropriate.

\paragraph{Compute.}

All canonical experiments were run on AWS EC2 \texttt{g5.12xlarge} instances, each with 4 NVIDIA A10G GPUs, 24GB GPU memory per GPU, 48 vCPUs, 192GiB CPU memory, and local NVMe storage. Experiments were launched through Hydra scripts with one training worker per GPU, so a fully utilized instance ran four independent training jobs concurrently. All models were implemented in PyTorch and trained with CUDA automatic mixed precision.

The main reported results required approximately 80--120 \texttt{g5.12xlarge} instance-hours, corresponding to roughly 320--480 A10G GPU-hours. This estimate includes the canonical runs for the four main architectures across the principal datasets and ID/OOD configurations used in the main paper. Including exploratory sweeps, failed runs, dataset preprocessing, additional seeds, larger policy-count experiments, and ablations over episode length, episode count, latent size, and dataset scale, the total compute used during the project was approximately 250--400 \texttt{g5.12xlarge} instance-hours, or about 1000--1600 A10G GPU-hours.

Run time varied substantially by dataset and model. Small synthetic GRF and public MuJoCo runs typically completed in minutes to under an hour per seed. Larger Hopper 10$\times$/20$\times$, Lichess, DROID, and FastF1 runs took longer because they used more episodes, longer sequences, or more policies. Behavioral INR also incurs extra inference-time cost because test representations are obtained by optimizing the episode latent while freezing the shared INR parameters. Per-run outputs include the composed Hydra configuration, metrics logs, evaluation summaries, and checkpoints, which will be released with the code.

\paragraph{Broader Impacts.}
Policy identification from behavior can raise privacy and security concerns when applied to human datasets. We use public or benchmark datasets and anonymized policy proxies where applicable, but deployment on sensitive human behavior should require consent, access controls, and privacy-preserving preprocessing.

\begin{table}[t]
\centering
\small
\caption{
Real-world OOD definitions. These domains are less controlled than synthetic GRF or checkpoint-generated MuJoCo, but they match the motivating setting: heterogeneous behavioral data without training-time policy labels.
}
\label{tab:real_world_ood_definitions}
\begin{tabular}{p{0.18\columnwidth}p{0.22\columnwidth}p{0.45\columnwidth}}
\toprule
Domain & Policy proxy & OOD definition \\
\midrule
Lichess & Player & Held-out game contexts that change the board-position distribution; the model must identify player-specific move choice rather than memorizing openings. \\
DROID & Collector / demonstrator & Held-out manipulation task family; in the balanced low-dimensional split, \texttt{remove} is held out. \\
FastF1 & Driver & Held-out racing contexts, including circuit- or session-conditioned splits; the model must identify driver behavior from telemetry rather than track-specific states. \\
\bottomrule
\end{tabular}
\end{table}

\begin{figure*}[t]
    \centering
    \setlength{\tabcolsep}{2pt}

    \begin{tabular}{cc}

    \begin{subfigure}[b]{0.48\textwidth}
        \centering
        \includegraphics[width=0.48\linewidth]{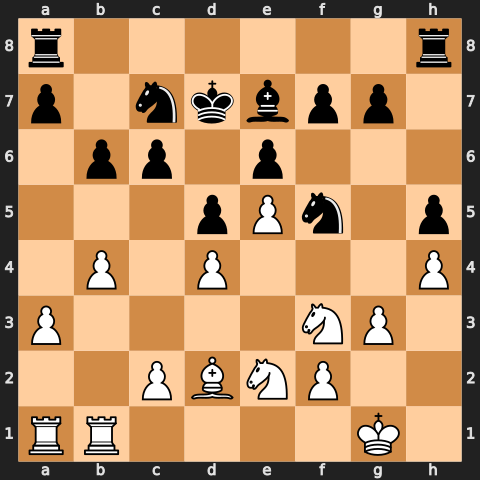}
        \hfill
        \includegraphics[width=0.48\linewidth]{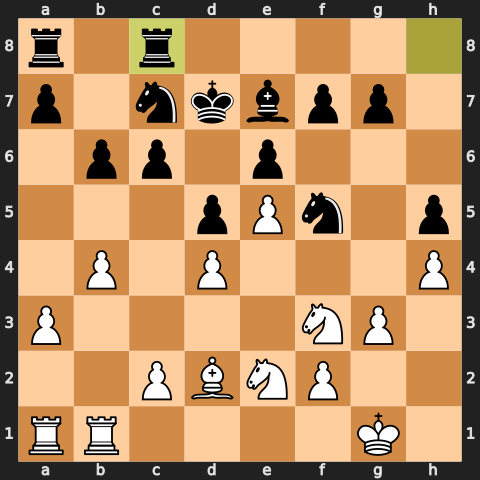}
        \caption{ID, lance5500 state sequence.}
    \end{subfigure}
    &
    \begin{subfigure}[b]{0.48\textwidth}
        \centering
        \includegraphics[width=0.48\linewidth]{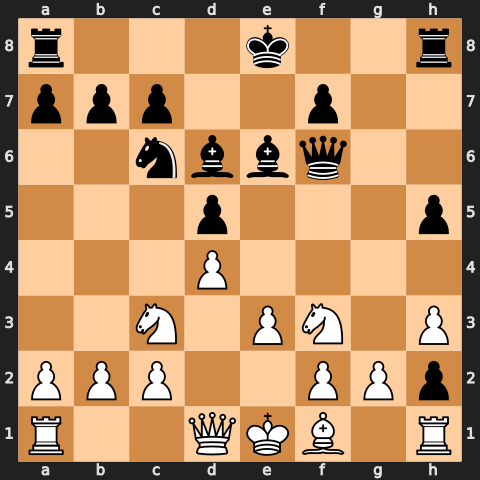}
        \hfill
        \includegraphics[width=0.48\linewidth]{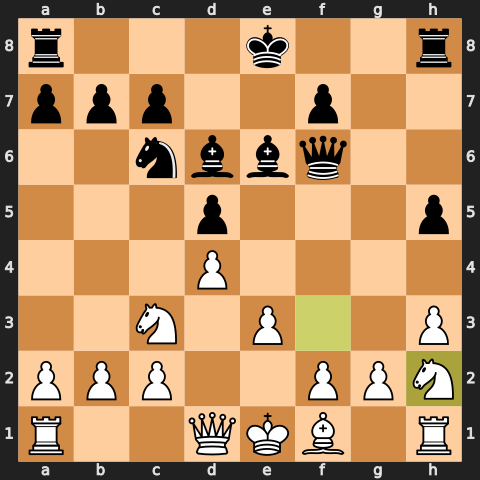}
        \caption{ID, penguingim1 state sequence.}
    \end{subfigure}
    \\[6pt]

    \begin{subfigure}[b]{0.48\textwidth}
        \centering
        \includegraphics[width=0.48\linewidth]{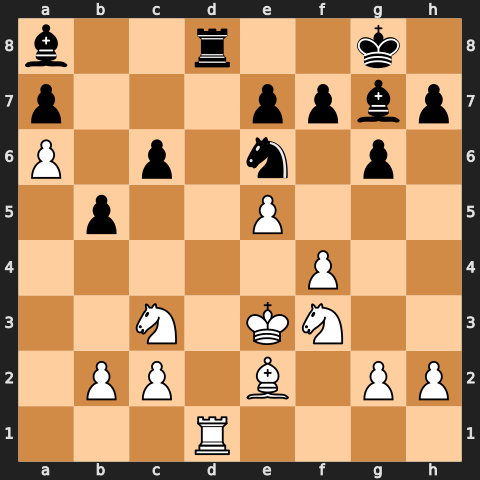}
        \hfill
        \includegraphics[width=0.48\linewidth]{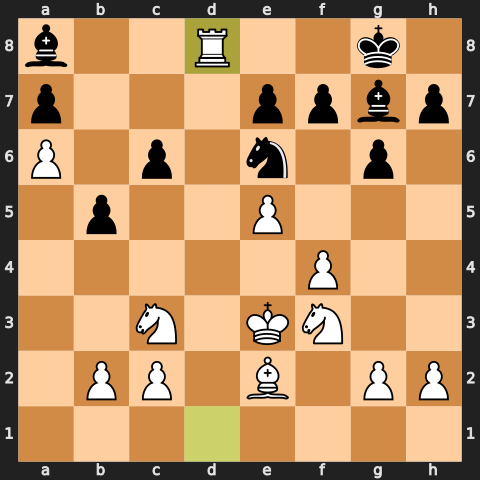}
        \caption{OOD, lance5500 state sequence.}
    \end{subfigure}
    &
    \begin{subfigure}[b]{0.48\textwidth}
        \centering
        \includegraphics[width=0.48\linewidth]{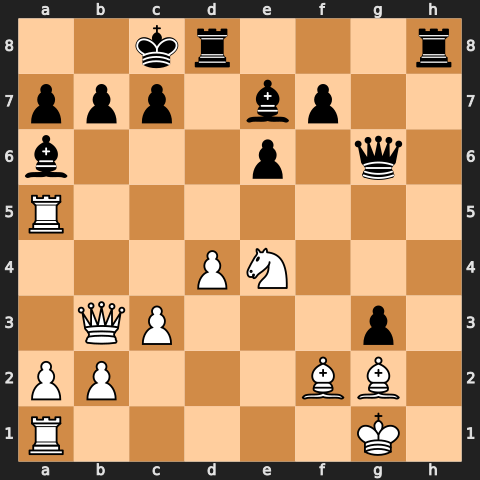}
        \hfill
        \includegraphics[width=0.48\linewidth]{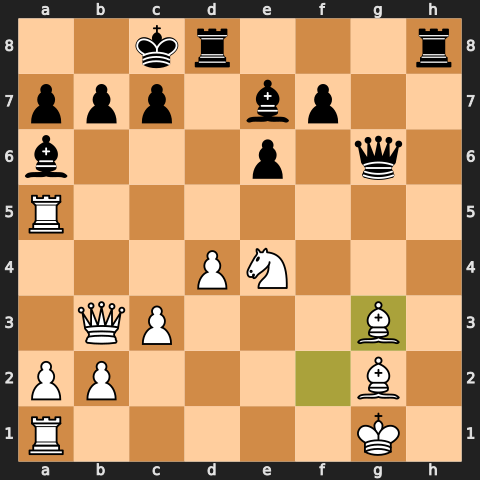}
        \caption{OOD, penguinim1 state sequence.}
    \end{subfigure}

    \end{tabular}

    \caption{
    Chess sequences from the Lichess dataset. PGN files store full chess games, while UCI denotes the standardized move notation used as the action label, e.g., \texttt{e2e4}. We construct OOD sequences by keeping only the tracked player's moves, representing each state as the board before that move, and holding out shared board-state regions identified by cross-player nearest-neighbor overlap. This tests whether the representation identifies a player from state-conditioned move choice rather than openings or board-state shortcuts.
    }
    \label{fig:chess}
\end{figure*}

\begin{figure*}[t]
\centering
\begin{subfigure}[b]{0.48\textwidth}
\centering
 \includegraphics[width=\linewidth]{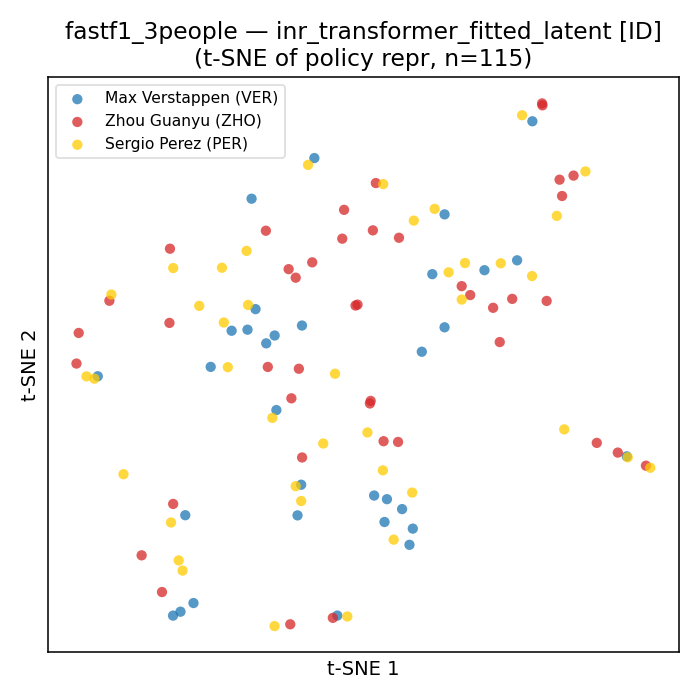}
\caption{Behavioral INR colored by player identity.}
\label{fig:f1_tracks_original}
\end{subfigure}
\hfill
\begin{subfigure}[b]{0.48\textwidth}
 \centering
\includegraphics[width=\linewidth]{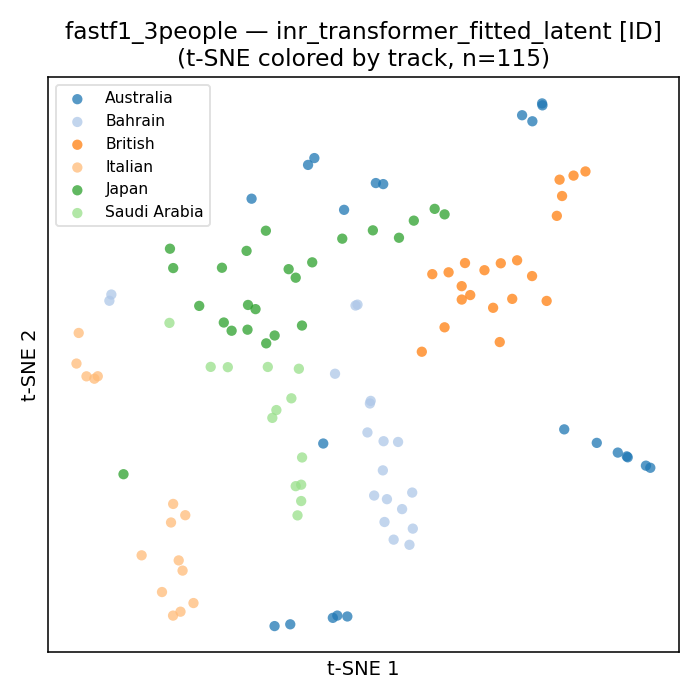}
\caption{Behavioral INR colored by track identity. }
\label{fig:f1_tracks_by_track}
\end{subfigure}
\caption{
Our embeddings hold multidimensional information; in Formula One, the policy is defined not only by the player but by the track. In fact, there exist more shortcuts to extracting track information than player identity, as shown here.
}
 \label{fig:f1_tracks}
\end{figure*}

\newpage
\begin{table}[t]
\centering
\small
\caption{
ID/OOD configurations. Each entry is policy: train $\rightarrow$ test; $\varnothing$ denotes an unseen training policy.
}
\label{tab:compExperiments}
\begin{tabular}{ll}
\toprule
Configuration & Policy splits \\
\midrule
No shift & $1:\mathrm{ID}\to\mathrm{ID}$, $2:\mathrm{ID}\to\mathrm{ID}$ \\
New policy & $1:\mathrm{ID}\to\mathrm{ID}$, $2:\mathrm{ID}\to\mathrm{ID}$, $3:\varnothing\to\mathrm{ID}$ \\
Single shift & $1:\mathrm{ID}\to\mathrm{OOD}$, $2:\mathrm{ID}\to\mathrm{ID}$ \\
Conflation & $1:\mathrm{OOD}\to\mathrm{OOD}$, $2:\mathrm{ID}\to\mathrm{ID}$ \\
Generalization & $1:\mathrm{OOD}\to\mathrm{ID}$, $2:\mathrm{OOD}\to\mathrm{ID}$ \\
Specialization & $1:\mathrm{ID}\to\mathrm{OOD}$, $2:\mathrm{ID}\to\mathrm{OOD}$ \\
Novel gen. & $1:\mathrm{OOD}\to\mathrm{ID}$, $2:\mathrm{OOD}\to\mathrm{ID}$, $3:\varnothing\to\mathrm{ID}$ \\
\bottomrule
\end{tabular}
\end{table}

\begin{table}[t]
\centering
\scriptsize
\caption{
DM Lab Seek-Avoid full split results. Values are means over seeds. Higher is better for P, K, and action accuracy; lower is better for NLL. Diffusion NLL is omitted because likelihood is not defined for that sampler.
}
\label{tab:dmlab_full_splits}
\resizebox{\columnwidth}{!}{%
\begin{tabular}{llcccc}
\toprule
Split & Model & P $\uparrow$ & K $\uparrow$ & Act. Acc. $\uparrow$ & NLL $\downarrow$ \\
\midrule
No shift & CVAE-Transf. & \textbf{1.000} & \textbf{1.000} & 0.590 & 1.079 \\
 & INR-Transf. & \textbf{1.000} & \textbf{1.000} & \textbf{0.644} & \textbf{0.956} \\
 & INR-Diff. & \textbf{1.000} & \textbf{1.000} & 0.260 & -- \\
 & \textbf{Ours} & \textbf{1.000} & \textbf{1.000} & 0.577 & 1.140 \\
\midrule
Single shift & CVAE-Transf. & 0.944 & 0.972 & 0.425 & 1.926 \\
 & INR-Transf. & \textbf{1.000} & \textbf{1.000} & \textbf{0.462} & 1.930 \\
 & INR-Diff. & \textbf{1.000} & \textbf{1.000} & 0.216 & -- \\
 & \textbf{Ours} & \textbf{1.000} & \textbf{1.000} & 0.403 & \textbf{1.878} \\
\midrule
Conflation & CVAE-Transf. & \textbf{1.000} & \textbf{1.000} & 0.468 & 1.593 \\
 & INR-Transf. & \textbf{1.000} & \textbf{1.000} & \textbf{0.499} & \textbf{1.510} \\
 & INR-Diff. & \textbf{1.000} & \textbf{1.000} & 0.192 & -- \\
 & \textbf{Ours} & \textbf{1.000} & \textbf{1.000} & 0.459 & 1.635 \\
\midrule
Generalization & CVAE-Transf. & \textbf{1.000} & \textbf{1.000} & \textbf{0.525} & 1.419 \\
 & INR-Transf. & \textbf{1.000} & \textbf{1.000} & 0.522 & \textbf{1.403} \\
 & INR-Diff. & \textbf{1.000} & \textbf{1.000} & 0.149 & -- \\
 & \textbf{Ours} & \textbf{1.000} & \textbf{1.000} & 0.517 & 1.501 \\
\midrule
Specialization & CVAE-Transf. & 0.972 & 0.972 & 0.295 & 2.678 \\
 & INR-Transf. & \textbf{1.000} & \textbf{1.000} & \textbf{0.317} & 2.798 \\
 & INR-Diff. & 0.889 & 0.972 & 0.172 & -- \\
 & \textbf{Ours} & \textbf{1.000} & \textbf{1.000} & 0.277 & \textbf{2.502} \\
\bottomrule
\end{tabular}%
}
\end{table}

\begin{table*}[t]
\centering
\scriptsize

\begin{minipage}[t]{0.52\textwidth}
\centering
\setlength{\tabcolsep}{3pt} 
\caption{
DM Lab Seek-Avoid under the \textsc{specialization} split. Values are mean $\pm$ standard deviation over seeds. Higher is better for probe, $k$NN, and action accuracy.
}
\label{tab:dmlab_results}
\begin{tabular}{lccc}
\toprule
Model & Probe Acc. $\uparrow$ & $k$NN-1 $\uparrow$ & Act. Acc. $\uparrow$\\
\midrule
CVAE-Transf. & $0.972 \pm 0.039$ & $0.972 \pm 0.039$ & $0.295 \pm 0.022$ \\
INR-Transf.  & $\mathbf{1.000 \pm 0.000}$ & $\mathbf{1.000 \pm 0.000}$ & $\mathbf{0.317 \pm 0.007}$ \\
INR-Diff.    & $0.889 \pm 0.079$ & $0.972 \pm 0.039$ & $0.172 \pm 0.003$  \\
\textbf{Ours} & $\mathbf{1.000 \pm 0.000}$ & $\mathbf{1.000 \pm 0.000}$ & $0.277 \pm 0.001$ \\
\bottomrule
\end{tabular}
\end{minipage}%
\hfill 
\begin{minipage}[t]{0.44\textwidth}
\centering
\setlength{\tabcolsep}{5pt} 
\caption{
FastF1 all-policy. Values are probe accuracies. Behavioral INR is best across the three listed splits, indicating that fitted functional representations scale better when the number of policies increases. The all-driver setting contains $21$ policy identities, so chance accuracy is $0.0476$.
}
\label{tab:fastf1_scaling}
\begin{tabular}{lcc}
\toprule
Model & \textsc{No-shift} & \textsc{Generalization} \\
\midrule
CVAE-Transf. &  0.143 & 0.043 \\
INR-Transf.  &  0.119 & 0.043 \\
INR-Diff.    &  0.071 & 0.048 \\
\textbf{Ours} &  \textbf{0.190} & \textbf{0.053} \\
\bottomrule
\end{tabular}
\end{minipage}

\end{table*}

\begin{table*}[t]
\centering
\scriptsize
\caption{
Full real-world all-policy representation results from the aggregate table. Each cell reports the mean over seeds. P is linear-probe accuracy and K is $k$NN-1 accuracy. Higher is better.
}
\label{tab:real_world_full_splits}
\resizebox{\textwidth}{!}{%
\begin{tabular}{llcccccccc}
\toprule
Domain & Split & CVAE-Transf. P & CVAE-Transf. K & INR Transf. P & INR Transf. K & INR Diff. P & INR Diff. K & Ours P & Ours K \\
\midrule
FastF1 & No shift & 0.240 & 0.082 & 0.067 & 0.019 & 0.101 & 0.053 & \textbf{0.274} & \textbf{0.111} \\
 & Single shift & 0.205 & 0.123 & 0.098 & 0.041 & 0.098 & 0.066 & \textbf{0.352} & \textbf{0.139} \\
 & Conflation & 0.189 & 0.115 & 0.164 & 0.082 & 0.205 & 0.049 & \textbf{0.295} & \textbf{0.131} \\
 & Generalization & 0.043 & 0.034 & 0.043 & 0.038 & 0.048 & 0.034 & \textbf{0.053} & \textbf{0.067} \\
 & Specialization & 0.143 & 0.000 & 0.119 & 0.000 & 0.071 & 0.000 & \textbf{0.191} & 0.000 \\
 & New policy & 0.197 & 0.082 & 0.106 & 0.062 & 0.106 & 0.067 & \textbf{0.245} & \textbf{0.096} \\
 & Novel gen. & 0.043 & \textbf{0.067} & \textbf{0.058} & 0.043 & 0.048 & 0.029 & 0.053 & \textbf{0.067} \\
\midrule
DROID & No shift & \textbf{0.700} & 0.567 & 0.656 & 0.433 & 0.678 & 0.522 & 0.622 & \textbf{0.600} \\
 & Single shift & \textbf{0.735} & 0.765 & 0.706 & 0.794 & 0.706 & 0.765 & 0.618 & \textbf{0.824} \\
 & Conflation & \textbf{0.912} & \textbf{0.794} & 0.853 & 0.765 & 0.765 & \textbf{0.794} & 0.588 & 0.706 \\
 & Generalization & \textbf{0.600} & \textbf{0.589} & 0.500 & 0.556 & 0.433 & 0.544 & 0.444 & 0.533 \\
 & Specialization & \textbf{0.667} & 0.000 & \textbf{0.667} & 0.000 & 0.500 & 0.000 & \textbf{0.667} & 0.000 \\
 & New policy & 0.578 & 0.533 & \textbf{0.600} & 0.422 & \textbf{0.600} & 0.456 & 0.511 & \textbf{0.600} \\
 & Novel gen. & \textbf{0.511} & \textbf{0.589} & 0.500 & 0.567 & 0.378 & 0.556 & 0.322 & 0.544 \\
\midrule
Lichess & No shift & \textbf{0.765} & \textbf{0.643} & 0.549 & 0.383 & 0.513 & 0.421 & 0.374 & 0.353 \\
 & Single shift & \textbf{0.741} & 0.641 & 0.531 & 0.676 & 0.542 & \textbf{0.708} & 0.257 & 0.567 \\
 & Conflation & \textbf{0.754} & 0.522 & 0.688 & 0.547 & 0.692 & \textbf{0.654} & 0.263 & 0.565 \\
 & Generalization & \textbf{0.512} & 0.389 & 0.451 & 0.402 & 0.417 & \textbf{0.440} & 0.330 & 0.420 \\
 & Specialization & \textbf{0.492} & \textbf{0.379} & 0.466 & 0.343 & 0.388 & 0.356 & 0.350 & 0.327 \\
 & New policy & \textbf{0.651} & \textbf{0.643} & 0.508 & 0.384 & 0.459 & 0.412 & 0.438 & 0.349 \\
 & Novel gen. & \textbf{0.473} & 0.392 & 0.412 & 0.405 & 0.366 & \textbf{0.428} & 0.331 & 0.420 \\
\bottomrule
\end{tabular}%
}
\end{table*}


\begin{table*}[t]
\centering
\scriptsize
\setlength{\tabcolsep}{3.5pt} 
\caption{
\textsc{Generalization} summary across the main datasets. Values are mean $\pm$ standard deviation over seeds. P is linear-probe accuracy; K is $k$NN-1 accuracy; NMSE and MedSE are continuous-action prediction losses; Acc. and NLL are discrete-action metrics. Missing metrics are not defined for that dataset in the aggregate.
}
\label{tab:generalization_summary_full}

\begin{tabular}{llcccccc}
\toprule
Domain & Model & P $\uparrow$ & K $\uparrow$ & NMSE $\downarrow$ & MedSE $\downarrow$ & Acc. $\uparrow$ & NLL $\downarrow$ \\
\midrule
Synthetic GRF & CVAE-Transf. & $0.567 \pm 0.141$ & $0.433 \pm 0.079$ & -- & -- & -- & -- \\
10x all policies & INR-Transf.  & $0.500 \pm 0.016$ & $0.333 \pm 0.031$ & -- & -- & -- & -- \\
                & INR-Diff.    & $0.544 \pm 0.016$ & $0.433 \pm 0.236$ & -- & -- & -- & -- \\
                & \textbf{Ours} & $\mathbf{0.611 \pm 0.110}$ & $\mathbf{0.478 \pm 0.173}$ & -- & -- & -- & -- \\
\midrule
Hopper 20x & CVAE-Transf. & $0.724 \pm 0.027$ & $0.994 \pm 0.009$ & $1.252 \pm 0.166$ & $0.392 \pm 0.043$ & -- & -- \\
           & INR-Transf.  & $0.603 \pm 0.145$ & $0.987 \pm 0.018$ & $0.995 \pm 0.043$ & $0.400 \pm 0.029$ & -- & -- \\
           & INR-Diff.    & $0.660 \pm 0.045$ & $\mathbf{1.000 \pm 0.000}$ & $\mathbf{0.968 \pm 0.016}$ & $\mathbf{0.382 \pm 0.005}$ & -- & -- \\
           & \textbf{Ours} & $\mathbf{0.744 \pm 0.000}$ & $\mathbf{1.000 \pm 0.000}$ & $1.105 \pm 0.082$ & $0.464 \pm 0.038$ & -- & -- \\
\midrule
DMLab & CVAE-Transf. & $\mathbf{1.000 \pm 0.000}$ & $\mathbf{1.000 \pm 0.000}$ & -- & -- & $\mathbf{0.525 \pm 0.004}$ & $1.419 \pm 0.032$ \\
Seek-Avoid & INR-Transf.  & $\mathbf{1.000 \pm 0.000}$ & $\mathbf{1.000 \pm 0.000}$ & -- & -- & $0.522 \pm 0.000$ & $\mathbf{1.403 \pm 0.016}$ \\
           & INR-Diff.    & $\mathbf{1.000 \pm 0.000}$ & $\mathbf{1.000 \pm 0.000}$ & -- & -- & $0.149 \pm 0.010$ & -- \\
           & \textbf{Ours} & $\mathbf{1.000 \pm 0.000}$ & $\mathbf{1.000 \pm 0.000}$ & -- & -- & $0.517 \pm 0.011$ & $1.501 \pm 0.011$ \\
\midrule
FastF1 & CVAE-Transf. & $0.043 \pm 0.007$ & $0.034 \pm 0.007$ & -- & -- & -- & -- \\
all drivers & INR-Transf.  & $0.043 \pm 0.007$ & $0.038 \pm 0.000$ & -- & -- & -- & -- \\
            & INR-Diff.    & $0.048 \pm 0.014$ & $0.034 \pm 0.034$ & -- & -- & -- & -- \\
            & \textbf{Ours} & $\mathbf{0.053 \pm 0.048}$ & $\mathbf{0.067 \pm 0.041}$ & -- & -- & -- & -- \\
\midrule
DROID & CVAE-Transf. & $\mathbf{0.600 \pm 0.063}$ & $\mathbf{0.589 \pm 0.141}$ & -- & -- & -- & -- \\
all policies & INR-Transf.  & $0.500 \pm 0.016$ & $0.556 \pm 0.189$ & -- & -- & -- & -- \\
             & INR-Diff.    & $0.433 \pm 0.079$ & $0.544 \pm 0.110$ & -- & -- & -- & -- \\
             & \textbf{Ours} & $0.444 \pm 0.126$ & $0.533 \pm 0.031$ & -- & -- & -- & -- \\
\midrule
Lichess & CVAE-Transf. & $\mathbf{0.512 \pm 0.000}$ & $0.389 \pm 0.000$ & -- & -- & -- & -- \\
all policies & INR-Transf.  & $0.451 \pm 0.000$ & $0.402 \pm 0.000$ & -- & -- & -- & -- \\
             & INR-Diff.    & $0.417 \pm 0.000$ & $\mathbf{0.440 \pm 0.000}$ & -- & -- & -- & -- \\
             & \textbf{Ours} & $0.330 \pm 0.000$ & $0.420 \pm 0.000$ & -- & -- & -- & -- \\
\bottomrule
\end{tabular}

\end{table*}

\end{document}